\documentclass[11pt,a4paper]{article}
\usepackage[hyperref]{emnlp2018}
\usepackage{times}
\usepackage{latexsym}

\usepackage{url}
\usepackage{graphics}

\aclfinalcopy 

\usepackage{lipsum}

\newcommand\blfootnote[1]{%
  \begingroup
  \renewcommand\thefootnote{}\footnote{#1}%
  \addtocounter{footnote}{-1}%
  \endgroup
}

\title{
Retrieve and Refine: \\ Improved Sequence Generation Models For Dialogue}

\author{Jason Weston, Emily Dinan and Alexander H. Miller \\
  Facebook AI Research \\
  {\tt jase@fb.com, edinan@fb.com, ahm@fb.com} 
  \\}

\date{}

\begin{document}
\maketitle
\begin{abstract}
  Sequence generation models for dialogue are known 
  to have several  problems: they tend
  to produce short, generic sentences that are uninformative and unengaging. Retrieval models on the other hand can surface interesting responses, 
  but are restricted to the given retrieval set leading to erroneous replies that cannot be tuned to the specific context.
  In this work 
  we develop a model that combines the two approaches
  to avoid both their deficiencies: 
  first retrieve a response and then refine it --
  the final sequence generator treating the retrieval as additional context.
  We show on the recent 
  {\sc ConvAI2} challenge task our approach produces responses superior to both standard retrieval and generation models in human evaluations.
\end{abstract}\blfootnote{Proceedings of the 2018 EMNLP Workshop SCAI: The 2nd International Workshop on Search-Oriented Conversational AI 978-1-948087-75-9}

\section{Introduction}

Sequence generation models like Seq2Seq \citep{sutskever2014sequence}
are increasingly popular for tasks such as machine translation (MT)
and summarization, where generation is suitably constrained by the source sentence. However, obtaining good performance on dialogue tasks, where the context still allows many 
interpretations, remains
an open problem despite much recent work
\cite{serban2016generative}.
Several authors report the issue that they produce short, generic sentences containing frequent words -- the so-called ``I don't know'' problem -- as that response
  can work as a reply in many instances, but is uninformative and unengaging.
  Retrieval models \cite{ji2014information} do not have this problem, but instead either produce engaging responses 
 or else completely erroneous ones which they cannot tune   to the specific context, as they can only produce a valid reply if it  is in the retrieval set.
  
  In this work we propose a Retrieve and Refine model to gain the advantages of both methods, and avoid both their disadvantages. Models that produce an initial prediction and then refine it are growing in traction in NLP. 
  They have been used in MT and summarization either for  refinement of initial predictions \cite{junczys2017exploration,niehues2016pre,novak2016iterative,xia2017deliberation,grangier2017quickedit} or combining with retrieval \cite{gu2017search,cao2018retrieve}, as well as for 
  sentence correction or refinement without context
  \cite{guu2017generating,schmaltz2017adapting}.
  There is little work in applying these methods to dialogue;
   one work we are aware of has been done concurrently with ours is  \citet{pandey2018exemplar}.
  The usefulness of our approach is shown with
  detailed  experiments on the ConvAI2 dataset\footnote{\url{http://convai.io/}} 
  which is a chit-chat task to get to know the other speaker's profile,
  obtaining generations superior to both retrieval and sequence generation models in human evaluations.
  

\section{Retrieve and Refine}

The model we propose in this work is remarkably 
straight-forward:
we take a standard generative model and concatenate the output of a retrieval model  to its usual input, and then generate as usual, training the model under this setting.
%

For the generator, we use a standard Seq2Seq model:
a 2-layer LSTM with attention. 
For the retriever, we use the Key-Value Memory Network 
\cite{miller2016key}
already shown to perform well for this dataset \cite{zhang2018personalizing}, which attends over the dialogue history, to learn input and candidate retrieval embeddings that match using cosine similarity.
The top scoring utterance
is provided  as input to our Seq2Seq model in order to refine it, prepended with a special separator token. For both models we use the code available in ParlAI\footnote{\small\url{http://parl.ai}}. At test time the retriever retrieves candidates from the training set.

To train our model we first precompute the retrieval result for every dialogue turn in the training set, but instead of using the top ranking results we
 rerank the top 100 predictions of each by their similarity to the label (in embedding space). 
Following  \citet{guu2017generating}
this should help avoid the problem of the refinement being too far away from the original retrieval. 
We then append the chosen utterances to the input sequences used to train Seq2Seq.
 We refer to our model as
{\em{RetrieveNRefine}}, or  {\em{RetNRef}} for short.
We also consider two variants of the model in the following that we found improve the results.

\paragraph{Use Retriever More} 
In our vanilla model, we noticed there was not enough attention being paid to the retrieval utterance by the generator.  As the input to Seq2Seq is the dialogue history  concatenated with the retrieval utterance, truncating the history is one way to pay more attention to the retrieval.
In particular for the ConvAI2 dataset we clip the initial profile sentences at the start of the dialogue, forcing the model to more strongly rely on the retriever which still has them.\footnote{Architectural changes might also  deal with this issue, e.g. treating the two inputs as independent sources to do attention over, but we take the simplest possible approach here.} 
We refer to this modification as {\em{RetrieveNRefine$^{+}$}}.

\paragraph{Fix Retrieval Copy Errors}

Our model learns to sometimes ignore the retrieval (when it is bad), sometimes use it partially, and other times simply copy it.
However, when it is mostly copied but only changes a word or two, we observed it made mistakes more often than not, leading to less meaningful utterances. We thus also consider a variant that exactly copies the retrieval if the model generates with large word overlap (we chose $>$60\%). 
Otherwise, we leave the generation untouched.\footnote{Other approaches might also 
help with this problem such as  using an explicit copy mechanism or to use BPE tokenization \cite{fan2017controllable}, but we leave those for future work.}
We refer to this as
{\em{RetrieveNRefine$^{++}$}}.

\section{Experiments}

We conduct experiments on the recent ConvAI2 challenge
dataset which
uses a modified version of the 
PersonaChat dataset \cite{zhang2018personalizing}  (larger, and with different processing). The dataset consists of conversations between crowdworkers who were randomly paired and asked to act the part of a given persona (randomly assigned from 1155 possible personas, created by another set of workers), chat naturally, and get to know each other during the conversation. 
 There are around 160,000 utterances in around 11,000 dialogues, with 2000 dialogues for validation and test, which use non-overlapping personas.

\subsection{Automatic Evaluation and Analysis}

\paragraph{Perplexity}
Dialogue is known to be notoriously hard to 
evaluate with automated metrics 
\cite{liu2016not}.
In contrast to machine translation, there is much 
less constraint on the output with many valid answers with little 
word overlap, e.g. there are many answers to ``what are you doing tonight?''.
Nevertheless many recent papers report perplexity results
in addition to human judgments.
For the retrieve and refine case, perplexity evaluation is particularly flawed: if the retrieval points the model to a response that is very different from (but equally valid as) the true response, the model might focus on refining that and get poor perplexity.


\begin{table}[t!]
\begin{center}
\begin{tabular}{l|r}
\hline
RetNRef Retrieval Method &  PPL  \\ 
\hline
None (Vanilla Seq2Seq)          & 31.4  \\
Random label                    & 32.0  \\
Memory Network         & 31.8 \\
True label's neighbor & 25.9 \\
True label                      & 9.2  \\
\hline
\end{tabular}
\end{center}
\caption{Perplexity on the ConvAI2 task test set with different types of retriever for RetNRef, see text.
\label{table:ppl_res}
}
\end{table}

\begin{table}[t!]
\begin{center}
\begin{tabular}{l|l|l|l|l}
\hline
\bf        & Word & Char &  \multicolumn{2}{c}{Rare Word \%}  \\ 
 Method & ~cnt  & ~cnt &  $<$100 & $<$1k  \\ 
\hline
Seq2Seq          & 11.7 & 40.5  & 0.4\%& 5.8\% \\ 
RetNRef          & 11.8 & 40.4  & 1.1\%& 6.9\% \\
RetNRef$^{+}$    & 12.1  &45.0  & 1.7\% &  10.1\% \\ 
RetNRef$^{++}$   & 12.7 & 48.1 & 2.3\% & 10.9\% \\
MemNet      & 13.1 & 54.5  &  4.0\% & 15.3\% \\
Human         & 13.0 & 54.6  & 3.0\% & 11.5\% \\
\hline
\end{tabular}
\end{center}
\caption{Output sequence statistics for the methods. Seq2Seq generates shorter sentences with more common words than humans, which RetNRef alleviates.
\label{table:stats}
}
\end{table}
\begin{table}[t!]
\begin{center}
\begin{tabular}{l|l|l|l|l}
\hline
Method  & {\small $<$30\%} & {\small 30-60\%} & {\small 60-80\%} & {\small $>$80\%}  \\ 
\hline
Seq2Seq        & 56\% & 34\% & 7\%  & 3\% \\
RetNRef        & 41\% & 38\% & 13\%  & 8\% \\ 
RetNRef$^{+}$  & 26\% & 20\% & 12\%  & 42\% \\
RetNRef$^{++}$ & 26\% & 20\% & 0\%   & 53\% \\ 
\hline
\end{tabular}
\end{center}
\caption{
Word overlap between retrieved and generated utterances in RetNRef, and between Seq2Seq and the Memory Network retriever (first row). 
\label{table:bin-stats}
}
\end{table}

\begin{table*}[t]
  \centering
  \begin{tabular}{l|cccc}
  \hline
  {\bf Method} & \textbf{Engagingness} & \textbf{Fluency} &  \textbf{Consistency} & 
   \textbf{Persona}   \\
  \hline
  Seq2Seq (PPL)        &  2.70(1.17) & 3.50(1.37) & 3.90(1.37) & 0.90(0.29)\\
  Seq2Seq (100 epochs) &  2.76(1.15) & 3.53(1.14) & 3.84(1.38) & 0.85(0.35) \\
  Memory Network       &   3.66(1.26) & 3.83(1.26) & 3.61(1.36) & 0.73(0.44) \\
  RetrieveNRefine       &   2.94(1.26) & 3.65(1.28) & 3.72(1.32) & 0.90(0.30) \\
  RetrieveNRefine$^{+}$ &3.50(1.33) & 3.63(1.13) & 3.55(1.33) & 0.71(0.45) \\
  RetrieveNRefine$^{++}$&3.80(1.18) & 3.74(1.19) & 3.80(1.40) & 0.65(0.47)  \\
  \hline
  \end{tabular}
 \caption{{Human Evaluation} scores for the models,scoring fluency, engagingness, consistency and persona detection,  with standard deviation in parentheses.
We consider engagingness  to be the most important metric.
   \label{tab:human-eval}
    }
    \begin{center}
\vspace{3mm}
\begin{tabular}{l|lllll}
\hline 
Comparison (A vs. B)                  & Win Rate  & A Wins & B Wins& Tie & p-value\\
RetrieveNRefine$^{++}$ vs. Memory Network & 54.5\% &     340 & 284 & 572      & 0.027  \\ 
RetrieveNRefine$^{++}$ vs. Seq2Seq   & 53.7\%     & 571 & 492 & 203  & 0.016 \\
\hline
RetrieveNRefine$^{++}$(retrieved) vs.Seq2Seq & 53.8\% & 290 & 249 & 87 &  \\
RetrieveNRefine$^{++}$(generated) vs.Seq2Seq & 53.6\% & 281 & 243 & 116 & \\
\hline
RetrieveNRefine$^{+}$  vs. Memory Network & 51.63\% &  492 & 461 & 243 & \\ 
RetrieveNRefine$^{+}$ vs. Human              & 30.13\% & 69 & 160 & 14 & \\
Seq2Seq  vs. Human              & 26.84\% & 62 & 169 & 22 & \\
\end{tabular}
\vspace{-1mm}
\caption{ A/B testing by humans comparing  model pairs, reporting win rates for A over B (excluding ties).
\label{table:pairfight}
}
\end{center}
\end{table*}

We therefore test our model by considering various types of retrieval methods:
(i) the best performing existing retriever model, the Memory Network approach from \citet{zhang2018personalizing} (retrieving from the training set),
(ii) a retriever that returns a random utterance from the training set,
(iii) the true label given in the test set, and (iv)
the closest nearest neighbor from the training set utterances to the true label, as measured by the embedding space of the Memory Network retriever model. 
While (iii) and (iv) cannot be used  in a deployed system as they are unknown, they can be used as a sanity check: a useful retrieve and refine should improve perplexity if given these as input.
We also compare to a standard Seq2Seq model, i.e. no retrieval.

The results are given in Table 
\ref{table:ppl_res}. They show that the RetNRef model 
can indeed improve perplexity with label neighbors or the label itself. However, surprisingly there is almost no difference between using no retrieval, random labels or our best retriever. 
The RetNRef$^{++}$ model -- that truncates the dialogue history and focuses more on the retrieval utterance -- does even worse in terms of perplexity: 48.4 using the Memory Network retriever.
However, poor perplexity does not mean human judgments of the generated sequences will not improve; in fact we will see that they do in the next section.
How to automatically evaluate these kinds of models still remains an open problem. 

\paragraph{Word Statistics}

Another way to measure the salience of a generation model is to compare it to human utterances in terms of word statistics.
We analyze the word statistics of our models in Table \ref{table:stats}. Seq2Seq models are known to produce short sentences with more common words than humans. 
The statistics on the ConvAI2 dataset bear this out, where
the Seq2Seq model responses have lower word and character counts and use fewer rare words than the human responses.
The RetNRef model (using the Memory Network retriever, retrieving from the training set) makes some improvements in this regard, e.g. doubling the use of rare words (with frequency less than 100) and smaller gains for words with frequency less than 1000,
but are still not close to human statistics. 
The RetNRef$^{++}$ model which boosts the use of the retrieval does better in this regard, making the statistics much closer to human ones. Of course these metrics do not measure whether the utterances
are semantically coherent, but it is encouraging to see a model using rare words as without this we believe it is hard for it to be engaging.

Table \ref{table:bin-stats} compares the word overlap between retrieved and generated sentences in the RetNRef variants in order to measure if RetNRef is either ignoring the retriever, or else paying too much attention to and copying it. As comparison, the first row also shows the overlap between the retriever and vanilla Seq2Seq which does not retrieve at all. The results show that RetNRef$^{++}$ has $>$80\% word overlap with the retriever output around half (53\%) of the time, whereas Seq2Seq and RetNRef very rarely overlap with the retriever (3\% and 8\% of the time respectively have $>$80\% overlap). This shows that our improved model RetNRef$^{++}$ does use the retriever, but can also generate novel content when it wants to, which a standard retriever cannot.

\begin{table*}
\begin{center}
\small
\resizebox{6.25in}{!}{
\begin{tabular}{l|l|||l}
\hline 
Person 1&hey , george here . lets be friends &
yeah. well, i wish you luck there. we bought a cute 2 story house.\\
Person 2 &hi george , i'm regina . do you enjoy dance ? &
nice , i don't think i have ever left ohio.\\
\hline
Seq2Seq& i do not but i do have a lot of time to dance &
i am sorry. i am so scared of twins.\\
MemNet& hello george my name is pink &
i am from mississippi . just moved back here from austin texas\\
RetNRef$^{+}$& i do. i enjoy dance. &
i am from tennessee. just moved here from austin texas.\\
\hline
\multicolumn{3}{c}{} \\
\hline 
Person 1&drinking coffee? what time is it there?  &
thank you for that , you are what you believe ?\\
Person 2 &830 pm haha my dad is a construction worker  &
look energy isn't physical , only spiritual \\
& so he works really late nights   & \\
\hline
Seq2Seq& i am a big guy so i can help you &
i am not sure what that means\\
MemNet& yep . moonlight is amazing but northern  &
do you do any sports ? swimming helps me keep my energy up.\\
&  light grab you too & \\
RetNRef$^{+}$& oh wow, that is a long time &
do you do any sports ? swimming helps me keep my energy up\\
\hline
\end{tabular}
}
\end{center}
\vspace{-1mm}
\caption{ Example dialogues (left) where RetNRef outperforms MemNet, and (right) where RetNRef outperforms Seq2Seq -- by either paying attention to the retriever or not. The first two rows are the previous two dialogue turns by Person 1 \& 2, the following rows are the predictions for the next utterance of Person 1 by the various models.
\label{table:ex-dialogs}
}
\end{table*}

\subsection{Evaluation by Human Judgement Scores}

Following the protocol in \citet{zhang2018personalizing},
we asked humans to 
conduct short dialogues with our models (100 dialogues each of 10-20 turns, so 600 dialogues in total),  and then measure the engagingness, consistency, and fluency (all scored out of 5) as well as to try to detect the persona that the model is using, given the choice between that and a random persona.
Consistency measures the coherence of the dialogue, e.g. ``I have a dog'' followed by ``I have no pets'' is not consistent.

The results are given in Table \ref{tab:human-eval}.
They show engagingness scores 
superior to Seq2Seq for all RetNRef variants,
and with RetNRef$^{++}$ slightly outperforming
the retriever which it conditions on. Importantly however,
it maintains this performance whilst still being able to generate text which a retrieval model cannot.
It also performs well in the other metrics, although like the Memory Network model, it is weaker at using the persona than Seq2Seq. 
Seq2Seq is inferior to the Memory Network Retriever in terms of engagement, in line with previous results. We also tried overtraining the Seq2Seq for 100 epochs instead of early stopping by validation on perplexity as it may overfit training sentences and act more as a retriever, but this did not help.

Some example dialogues of the RetNRef$^{++}$ model performing well (as scored by the evaluators) are shown in Tables \ref{table:rnr1-example} and \ref{table:rnr2-example}. Longer sentences from the bot (person 2) with more nuanced entity information typically come from attending to the retriever, whereas the generator can also produce shorter replies independent of the retriever that fit the context well.
There are still issues however, such as repeated phrases by the generator, and some tendency to copy the speaking partner's phrases that could be improved.

\begin{table*}[t]
  \begin{center}
    \begin{small}
      \begin{tabular}{l|l}
        \textbf{Persona 1} & \textbf{Persona 2}\\
I am relatively tall.    &  I like to play baseball.\\
I like mexican food.     & I love eating seafood.\\ 
I drive a subaru legacy. & I enjoy learning new languages.\\
My dog's name is Drunky.&  I'm currently going for a marketing degree.\\
I can touch my toes with my nose.& I want to design ads in the future. \\
\multicolumn{2}{l}{ }\\
\multicolumn{2}{l}{[PERSON 1:] I'm driving to the mexican place tonight for dinner, but my Subaru is at the rental shop.}\\
\multicolumn{2}{l}{[PERSON 2:] Yummy. I love mexican food, but I never eat vegetables}\\
\multicolumn{2}{l}{[PERSON 1:] Oh, I do, I think that's why i'm so tall!}\\
\multicolumn{2}{l}{[PERSON 2:] how tall are you? I do not... I enjoy studying languages.} \\
\multicolumn{2}{l}{[PERSON 1:] I'm 6 foot, a lot taller than my dog, Drunky.}\\
\multicolumn{2}{l}{[PERSON 2:] I am sorry to hear that. I want to be 6 foot tall.}\\
\multicolumn{2}{l}{[PERSON 1:] Do you have any talents? I can reach my toes using my nose.}\\
\multicolumn{2}{l}{[PERSON 2:] Cool! I listen to audiobooks and speak one asian language. you like sushi? I do.}\\ 
\multicolumn{2}{l}{[PERSON 1:] Oh sushi is my next favorite, after Mexican. What kind of car do you have?}\\
\multicolumn{2}{l}{[PERSON 2:] I have a Nissan Sentra. I love it.}\\ 
      \end{tabular}
      \caption{Example dialog between a human (Person 1) and the RetNRef$^{++}$ model (Person 2). References to never eating vegetables, listening to audiobooks and speaking one asian language and owning a Nissan Sentra appeared in the conditioned retrieved sentences, improving the overall engaging quality of the dialogue.
 \label{table:rnr1-example}}

    \end{small}
  \end{center}
\end{table*}

\begin{table*}[t]
  \begin{center}
    \begin{small}
      \begin{tabular}{l|l}
        \textbf{Persona 1} & \textbf{Persona 2}\\
I am a lacto-ovo vegetarian.    &  I have brown hair.\\
I like to hike.     & I am Latina.\\ 
I have red hair. & My mother works for special needs clients..\\
My favorite drink is sprite.&  I am a healer.\\
I have a pet parrot named Tasha.& I love my family. \\
\multicolumn{2}{l}{ }\\
\multicolumn{2}{l}{[PERSON 1:] I dont know about you but I love sprite.  Its my favorite.}\\
\multicolumn{2}{l}{[PERSON 2:] I like dr pepper a lot, but sprite is my second favorite.}\\
\multicolumn{2}{l}{[PERSON 1:] Sprites great!  I take it with me every day on my hikes.}\\
\multicolumn{2}{l}{[PERSON 2:] I bet. I love to go to the water} \\
\multicolumn{2}{l}{[PERSON 1:] Ya that sounds like a great time.  I go with my pet parrot Tasha when I go}\\
\multicolumn{2}{l}{[PERSON 2:] that is a good idea. I wish I had a parrot.}\\
\multicolumn{2}{l}{[PERSON 1:] Ya its fun.  As a pet owner you learn to love animals; im even a lactoovo vegetarian now}\\
\multicolumn{2}{l}{[PERSON 2:] I love animals. I am vegan so I eat vegan food.}\\ 
\multicolumn{2}{l}{[PERSON 1:] Sounds like me :D  im vegan, im ginger, and i love sprite}\\
\multicolumn{2}{l}{[PERSON 2:] I love animals. I am vegan so I eat a lot of it.}\\ 
      \end{tabular}
      \caption{Example dialog between a human (Person 1) and the RetNRef$^{++}$ model (Person 2).  References to Dr. Pepper and wishing to have a parrot appeared in the conditioned retrieved sentences, improving the overall engaging quality of the dialogue. 
There are still issues however, such as repeated phrases by the generator, failure to use the persona adequately, and some tendency to copy the speaking partner's phrases that could be improved.
 \label{table:rnr2-example}}
    \end{small}
  \end{center}
\end{table*}

\subsection{A/B Testing by Humans}

To compute a statistically more meaningful test,
and to evaluate models
more clearly against each other, we also tested
models by pairing them against each other (different to the protocol previously used for this dataset). 
We took human-human paired dialogues from the test set, up to a randomly chosen 
turn, and then computed the response of two rival models. The resulting dialogue history and the two choices are given to a third annotator to evaluate and asked which model is best; they can also choose ``unsure''. We collected over 1000 such examples, with each conversation being evaluated twice. 
The results are given in Table \ref{table:pairfight}.

RetrieveNRefine obtains statistically significant wins
over the retriever Memory Network model and the generator Seq2Seq model using a binomial two-tailed test, 
with win rates $\sim$54\%.
Breaking down the wins between when RetNRef$^{++}$ 
exactly copies the retrieval utterance vs. generates we see
that it chooses them about equally, with wins about equal in both cases. This shows it can effectively learn  when to choose the retrieval utterance (when it is good), and when to ignore it and generate instead (when it is bad). Table \ref{table:ex-dialogs}, which shows example outputs of our model, illustrates this.

RetNRef$^{+}$ sometimes loses out when making small changes to the retrieved text, for example it made changes to ``i once broke my nose trying to peak in on a jazz concert !'' by replacing {\em peak} with {\em glacier}. Recall that RetNRef$^{++}$ fixes this problem by exactly copying the retrieved text when there is insignificant word overlap with the generated text; as such, it has a correspondingly larger win rate against Memory Networks (54.5\% versus 51.63\%).

%

We also computed a small sample of A/B tests directly against humans rather than models, and again see the win rate is higher for RetNRef.

\section{Conclusion}

In conclusion, we showed that retrieval models can 
be successfully used to improve generation models in
dialogue, helping them avoid  common 
issues such as producing short sentences with frequent words that ultimately are not engaging. Our 
RetNRef$^{++}$ model has similar statistics to human utterances and provides more engaging conversations according to human judgments.

Future work should investigate improved ways to incorporate retrieval in generation, both avoiding the 
heuristics we used here to improve performance, 
and seeing if more sophisticated approaches than concatenation plus attention improve the results, for 
example by more clearly treating the inputs as independent
sources, or training the models jointly.

\bibliography{emnlp2018}

\begin{thebibliography}{17}
\expandafter\ifx\csname natexlab\endcsname\relax\def\natexlab#1{#1}\fi

\bibitem[{Cao et~al.(2018)Cao, Li, Li, and Wei}]{cao2018retrieve}
Ziqiang Cao, Wenjie Li, Sujian Li, and Furu Wei. 2018.
\newblock Retrieve, rerank and rewrite: Soft template based neural
  summarization.
\newblock In \emph{Proceedings of the 56th Annual Meeting of the Association
  for Computational Linguistics (Volume 1: Long Papers)}, volume~1, pages
  152--161.

\bibitem[{Fan et~al.(2017)Fan, Grangier, and Auli}]{fan2017controllable}
Angela Fan, David Grangier, and Michael Auli. 2017.
\newblock Controllable abstractive summarization.
\newblock \emph{arXiv preprint arXiv:1711.05217}.

\bibitem[{Grangier and Auli(2017)}]{grangier2017quickedit}
David Grangier and Michael Auli. 2017.
\newblock Quickedit: Editing text \& translations via simple delete actions.
\newblock \emph{arXiv preprint arXiv:1711.04805}.

\bibitem[{Gu et~al.(2017)Gu, Wang, Cho, and Li}]{gu2017search}
Jiatao Gu, Yong Wang, Kyunghyun Cho, and Victor~OK Li. 2017.
\newblock Search engine guided non-parametric neural machine translation.
\newblock \emph{arXiv preprint arXiv:1705.07267}.

\bibitem[{Guu et~al.(2017)Guu, Hashimoto, Oren, and Liang}]{guu2017generating}
Kelvin Guu, Tatsunori~B Hashimoto, Yonatan Oren, and Percy Liang. 2017.
\newblock Generating sentences by editing prototypes.
\newblock \emph{arXiv preprint arXiv:1709.08878}.

\bibitem[{Ji et~al.(2014)Ji, Lu, and Li}]{ji2014information}
Zongcheng Ji, Zhengdong Lu, and Hang Li. 2014.
\newblock An information retrieval approach to short text conversation.
\newblock \emph{arXiv preprint arXiv:1408.6988}.

\bibitem[{Junczys-Dowmunt and Grundkiewicz(2017)}]{junczys2017exploration}
Marcin Junczys-Dowmunt and Roman Grundkiewicz. 2017.
\newblock An exploration of neural sequence-to-sequence architectures for
  automatic post-editing.
\newblock \emph{arXiv preprint arXiv:1706.04138}.

\bibitem[{Liu et~al.(2016)Liu, Lowe, Serban, Noseworthy, Charlin, and
  Pineau}]{liu2016not}
Chia-Wei Liu, Ryan Lowe, Iulian~V Serban, Michael Noseworthy, Laurent Charlin,
  and Joelle Pineau. 2016.
\newblock How not to evaluate your dialogue system: An empirical study of
  unsupervised evaluation metrics for dialogue response generation.
\newblock \emph{arXiv preprint arXiv:1603.08023}.

\bibitem[{Miller et~al.(2016)Miller, Fisch, Dodge, Karimi, Bordes, and
  Weston}]{miller2016key}
Alexander Miller, Adam Fisch, Jesse Dodge, Amir-Hossein Karimi, Antoine Bordes,
  and Jason Weston. 2016.
\newblock Key-value memory networks for directly reading documents.
\newblock \emph{arXiv preprint arXiv:1606.03126}.

\bibitem[{Niehues et~al.(2016)Niehues, Cho, Ha, and Waibel}]{niehues2016pre}
Jan Niehues, Eunah Cho, Thanh-Le Ha, and Alex Waibel. 2016.
\newblock Pre-translation for neural machine translation.
\newblock \emph{arXiv preprint arXiv:1610.05243}.

\bibitem[{Novak et~al.(2016)Novak, Auli, and Grangier}]{novak2016iterative}
Roman Novak, Michael Auli, and David Grangier. 2016.
\newblock Iterative refinement for machine translation.
\newblock \emph{arXiv preprint arXiv:1610.06602}.

\bibitem[{Pandey et~al.(2018)Pandey, Contractor, Kumar, and
  Joshi}]{pandey2018exemplar}
Gaurav Pandey, Danish Contractor, Vineet Kumar, and Sachindra Joshi. 2018.
\newblock Exemplar encoder-decoder for neural conversation generation.
\newblock In \emph{Proceedings of the 56th Annual Meeting of the Association
  for Computational Linguistics (Volume 1: Long Papers)}, volume~1, pages
  1329--1338.

\bibitem[{Schmaltz et~al.(2017)Schmaltz, Kim, Rush, and
  Shieber}]{schmaltz2017adapting}
Allen Schmaltz, Yoon Kim, Alexander~M Rush, and Stuart~M Shieber. 2017.
\newblock Adapting sequence models for sentence correction.
\newblock \emph{arXiv preprint arXiv:1707.09067}.

\bibitem[{Serban et~al.(2016)Serban, Lowe, Charlin, and
  Pineau}]{serban2016generative}
Iulian~Vlad Serban, Ryan Lowe, Laurent Charlin, and Joelle Pineau. 2016.
\newblock Generative deep neural networks for dialogue: A short review.
\newblock \emph{arXiv preprint arXiv:1611.06216}.

\bibitem[{Sutskever et~al.(2014)Sutskever, Vinyals, and
  Le}]{sutskever2014sequence}
Ilya Sutskever, Oriol Vinyals, and Quoc~V Le. 2014.
\newblock Sequence to sequence learning with neural networks.
\newblock In \emph{Advances in neural information processing systems}, pages
  3104--3112.

\bibitem[{Xia et~al.(2017)Xia, Tian, Wu, Lin, Qin, Yu, and
  Liu}]{xia2017deliberation}
Yingce Xia, Fei Tian, Lijun Wu, Jianxin Lin, Tao Qin, Nenghai Yu, and Tie-Yan
  Liu. 2017.
\newblock Deliberation networks: Sequence generation beyond one-pass decoding.
\newblock In \emph{Advances in Neural Information Processing Systems}, pages
  1782--1792.

\bibitem[{Zhang et~al.(2018)Zhang, Dinan, Urbanek, Szlam, Kiela, and
  Weston}]{zhang2018personalizing}
Saizheng Zhang, Emily Dinan, Jack Urbanek, Arthur Szlam, Douwe Kiela, and Jason
  Weston. 2018.
\newblock Personalizing dialogue agents: I have a dog, do you have pets too?
\newblock \emph{arXiv preprint arXiv:1801.07243}.

\end{thebibliography}
\bibliographystyle{acl_natbib_nourl}

\end{document}